\documentclass[conference,usletter]{IEEEtran}
\usepackage{times,amsmath,amssymb}
\usepackage{slashbox}
\usepackage{graphicx}
\usepackage{xcolor}
\usepackage{algorithm2e}
\usepackage{algorithmicx}
\usepackage{setspace}
\usepackage[export]{adjustbox}
\usepackage{multirow}
\usepackage[noend]{algpseudocode}
\usepackage{braket}
\usepackage{stfloats}
\newtheorem{defn}{Definition}
\begin{document}

\title{Improving Efficiency in Convolutional Neural Network with Multilinear Filters}
\author{\IEEEauthorblockN{Dat Thanh Tran\IEEEauthorrefmark{1}, Alexandros Iosifidis\IEEEauthorrefmark{2} and Moncef Gabbouj\IEEEauthorrefmark{1}}
\IEEEauthorblockA{\IEEEauthorrefmark{1}Laboratory of Signal Processing, Tampere University of Technology, Tampere, Finland\\
\IEEEauthorrefmark{2}Department of Engineering, Electrical \& Computer Engineering, Aarhus University, Aarhus, Denmark\\
Email:\{dat.tranthanh,moncef.gabbouj\}@tut.fi, alexandros.iosifidis@eng.au.dk}\\

}

\maketitle

\begin{abstract}
The excellent performance of deep neural networks has enabled us to solve several automatization problems, opening an era of autonomous devices. However, current deep net architectures are heavy with millions of parameters and require billions of floating point operations. Several works have been developed to compress a pre-trained deep network to reduce memory footprint and, possibly, computation. Instead of compressing a pre-trained network, in this work, we propose a generic neural network layer structure employing multilinear projection as the primary feature extractor. The proposed architecture requires several times less memory as compared to the traditional Convolutional Neural Networks (CNN), while inherits the similar design principles of a CNN. In addition, the proposed architecture is equipped with two computation schemes that enable computation reduction or scalability. Experimental results show the effectiveness of our compact projection that outperforms traditional CNN, while requiring far fewer parameters.
\end{abstract}

\section{Introduction}\label{S:Intro}
In recent years, deep neural network architectures have excelled in several application domains, ranging from machine vision \cite{girshick2014rich, redmon2016you, waris2017cnn}, natural language processing \cite{hinton2012deep, graves2013speech} to biomedical \cite{zabihi2016heart, an2014deep} and financial data analysis \cite{tsantekidis2017using, tsantekidis2017forecasting}. Of those important developments, Convolutional Neural Network (CNN) has evolved as a main workhorse in solving computer vision tasks nowadays. The architecture was originally developed in the 1990s for handwritten character recognition using only two convolutional layers \cite{lecun1998gradient}. Over the years, with the development of Graphical Processing Units (GPUs) and efficient implementation of convolution operation, the depth of CNNs has been increased to tackle more complicated problems. Nowadays, prominent architectures such as Residual Network (ResNet) \cite{he2016deep} or Google Inception \cite{szegedy2015going} with hundreds of layers have become saturated. Researchers started to wonder whether millions of parameters are essential to achieve such performance. In order to extend the benefit of such deep nets to embedded devices with limited computation power and memory, recent works have focused on reducing the memory footprint and computation of a pre-trained network, i.e. they apply network compression in the post-training stage. In fact, recent works have shown that traditional network architectures such as Alexnet, VGG or Inception are highly redundant structures \cite{han2015deep,guo2016dynamic,chen2015compressing,wen2016learning,gong2014compressing,lin2016fixed,tai2015convolutional,denton2014exploiting,jaderberg2014speeding,ioannou2015training}. For example, in \cite{han2015deep} a simple heuristic based on magnitude of the weights was employed to eliminate the connections in a pre-trained network, which achieved considerable amount of compression without hurting the performance much. Additionally, representing network parameters with low bitwidth numbers, like in \cite{hubara2016quantized,gysel2016hardware,zhou2017balanced}, has shown that the performance of a 32-bit network can be closely retained with only 4-bit representations. It should be noted that the two approaches are complementary to each other. In fact, a compression pipeline called ``Deep Compression" \cite{han2015deep} which consists of three compression procedures, i.e. weight pruning, weight quantization and Huffman-based weight encoding, achieved excellent compression performance on AlexNet and VGG-16 architectures.

Along pruning and quantization, low-rank approximation of both convolutional layers and fully connected layers was also employed to achieve computational speed up \cite{denil2013predicting,novikov2015tensorizing,lin2016towards}. Viewed as high-order tensors, convolutional layers were decomposed using traditional tensor decomposition methods, such as CP decomposition \cite{jaderberg2014speeding,denton2014exploiting,lebedev2014speeding} or Tucker decomposition \cite{kim2015compression}, and the convolution operation is approximated by applying consecutive 1D convolutions.

Overall, efforts to remove redundancy in already trained neural networks have shown promising results by determining networks with a much simpler structure. The results naturally pose the following question: \textit{why should we compress an already trained network and not seek for a compact network representation that can be trained from scratch?}. Subsequently, one could of course exploit the above mentioned compression techniques to further decrease the cost. Under this perspective, the works in \cite{tai2015convolutional,ioannou2015training} utilizing a low-rank approximation approach were among the first to report simplified network structures.

The success of Convolutional Neural Networks can be attributed to four important design principles: sparse connectivity, parameter sharing, pooling and multilayer structure. Sparse connectivity (in convolutional layers) only allows local interaction between input neurons and output neurons. This design principle comes from the fact that in many natural data modalities such as images and videos local/neighboring values are often highly correlated. These groups of local values usually contain certain distinctive patterns, e.g. edges and color blobs, in images. Parameter sharing mechanism in CNNs enables the underlying model to learn location invariant cues. In other words, by sliding the filters over the input, the patterns can be detected regardless of the location. Pooling and multilayer structure design of deep neural networks in general and CNN in particular, captures the compositional hierarchies embedded within many natural signals. For example, in facial images, lower level cues such as edges, color and texture patterns form discriminative higher level cues of facial parts, like nose, eyes or lips. Similar compositional structure can be seen in speech or text, which are composed of phonemes, syllables, words and sentences. Although the particular structure of a deep network has evolved over time, the above important design principles remain unchanged. At the core of any convolution layer, each filter with $d
\times d \times C$ elements operates as a micro feature extractor that performs linear projection of each data patch/volume from a feature space of $d^2C$ dimensions to a real value. In order to enhance the discrimination power of this micro feature extractor, the authors of \cite{lin2013network} proposed to replace the GLM model by a general nonlinear function approximator, particularly the multilayer perceptron (MLP). The resulting architecture was dubbed Network in Network (NiN) since it consists of micro networks that perform the feature extractor functionality instead of simple linear projection.

In this paper, instead of seeking a more complex feature extractor, we propose to replace the linear projection of the traditional CNN by multilinear projection in the pursuit of simplicity. There has been a great effort to extend traditional linear methods to multilinear ones in an attempt to directly learn from the natural representation of the data as high order tensors \cite{li2014multilinear,yan2005discriminant,zhou2013tensor,thanh2017tensor,tao2007general,lu2008mpca}. The beauty of multilinear techniques lies in the property that the input tensor is projected simultaneously in each tensor mode, allowing only certain connections between the input dimensions and output dimensions, hence greatly reducing the number of parameters. Previous works on multilinear discriminant learning and multilinear regression \cite{guo2012tensor,li2014multilinear,yan2005discriminant,zhou2013tensor,thanh2017tensor,tao2007general} have shown competitive results of multilinear-based techniques. The proposed architecture still inherits the four fundamental design properties of a traditional deep network while utilizing multilinear projection as a generic feature extractor. The complexity of each feature extractor can be easily controlled through the ``rank" hyper-parameter. Besides a fast computation scheme when the network is compact, we also propose an alternative computation method that allows efficient computation when complexity increases.

The contribution of our paper can be summarized as follows:
\begin{itemize}
	\item We propose a generic feature extractor that performs multilinear mapping to replace the conventional linear filters in CNNs. The complexity of each individual feature extractor can be easily controlled via its rank, which is a hyperparameter of the method. By having the ability to adjust individual filter's complexity, the complexity of the entire network can be adjusted without the need of increasing the number of filters in a layer, i.e. the width of the layer. Since the proposed mapping is differentiable, the entire network can be easily trained end-to-end by using any gradient descent-based training process.
	\item We provide the analysis of computation and memory requirements of the proposed structure. In addition, based on the properties of the proposed mapping, we propose two efficient computation strategies leading to two different complexity settings.
	\item The theoretical analysis of the proposed approach is supported by experimental results in real-world classification problems, in comparison with CNN and the low-rank scheme in \cite{tai2015convolutional}.
\end{itemize}

The remainder of the paper is organized as follows: In section 2, we provide an overview of the related works focusing on designing compact network structures. Section 3 gives the necessary notations and definitions before presenting the proposed structure and its analysis. In section 4, we provide details of our experiment procedures, results and quantitative analysis. Section 5 concludes our work and discusses possible future extensions.

\section{Related Work}
Research focusing on the design of a less redundant network architecture has gained much more attention recently. One of the prominent design pattern is the \textit{bottleneck} unit which was first introduced in the ResNet architecture \cite{he2016deep}. The \textit{bottleneck} pattern is formed by two $1\times1$ convolution layers with some $3\times3$ convolution layers in between. The first $1\times1$ convolution layer is used to reduce the number of input feature maps while the latter is used to restore the number of output feature maps. Several works such as \cite{zhang2017shufflenet,wangdesign,szegedy2017inception} have incorporated the \textit{bottleneck} units into their network structure to reduce computation and memory consumed. Recently MobileNet architecture \cite{howard2017mobilenets} was proposed which replaced normal convolution operation by depthwise separable convolution layers. Constituted by depthwise convolution and pointwise convolution, the depthwise separable convolution layer performs the filtering and combining steps independently. The resulting structure is many times more efficient in terms of memory and computation. It should be noted that bottleneck design or depthwise separable convolution layer is a design on a macro level of the network structure in which the arrangements of layers are investigated to reduce computation.

On a micro level, the works in \cite{tai2015convolutional} and \cite{ioannou2015training} assumed a low rank structure of convolutional kernels in order to derive a compact network structure. In fact, low rank assumption has been incorporated into several designs prior to deep neural networks, such as dictionary learning, wavelet transform of high dimensional data. The first incorporation of low rank assumption in neural network compression was proposed in \cite{jaderberg2014speeding,lebedev2014speeding, denton2014exploiting}. In \cite{lebedev2014speeding}, CP decomposition was proposed to decompose the entire 4D convolutional layer into four 1D convolutions. Although the effective depth of the network remains the same, replacing one convolution operation by four can potentially lead to difficulty in training the network from scratch. With a carefully designed initialization scheme, the work of \cite{ioannou2015training} was able to train a mixture of low-rank filters from scratch with competitive performances. Improving on the idea of \cite{lebedev2014speeding}, a different low-rank structure that allows both approximating an already trained network and training from scratch was proposed in \cite{tai2015convolutional}. Specifically, let us denote a convolution layer of $N$ kernels by $\mathcal{W} \in \mathbb{R}^{d\times d\times C\times N}$, where $C$ and $d$ are the number of input feature maps and spatial size of the kernel, respectively. \cite{tai2015convolutional} proposed to approximate $\mathcal{W}$ using a vertical kernel $\mathcal{V} \in \mathbb{R}^{d\times 1\times C\times K}$ and a horizontal kernel $\mathcal{H} \in \mathbb{R}^{1\times d\times K\times N}$. The approximation is in the following form:

\begin{equation}\label{eq1}
\tilde{\mathcal{W}^{c}_{n}} \simeq \sum_{k=1}^{K}\mathcal{V}^{c}_{k}(\mathcal{H}^{k}_{n})^{T},
\end{equation}
where the superscript and subscript denote the index of the channel and the kernel respectively. $K$ is a hyper-parameter controlling the rank of the matrix approximation. Here $\mathcal{W}^{c}_{n}$ is just the 2D kernel weight of the $n$-th filter applied to the $c$-th channel of the input feature map; $\mathcal{V}^{c}_{k}$ and $\mathcal{H}^{k}_{n}$ are just $d$-dimensional vectors.

As can be seen from (\ref{eq1}), the authors simplify a convolutional layer by two types of parameter sharing. The first is the sharing of right singular vectors ($\mathcal{H}^{k}_{n}$) across all $C$ input channels within the $n$-th filter while the second enforces the sharing of left singular vectors ($\mathcal{V}^{c}_{k}$) across all $N$ filters. The work in \cite{tai2015convolutional} is closely related to ours since we avoid designing a particular initialization scheme by including a Batch Normalization step \cite{ioffe2015batch}. The resulting structure was easily trained from scratch with different network configurations.

\section{Proposed Method}\label{S:ProposedMethod}

We start this section by introducing some notations and definitions related to our work. We denote scalar values by either low-case or upper-case characters $(x, y, X, Y \dots)$, vectors by low-case bold-face characters $(\mathbf{x}, \mathbf{y}, \dots)$, matrices by upper-case bold-face characters $(\mathbf{A}, \mathbf{B}, \dots)$ and tensors by calligraphic capital characters $(\mathcal{X}, \mathcal{Y}, \dots)$. A tensor is a multilinear matrix with $K$ modes, and is defined as $\mathcal{X} \in \mathbb{R}^{I_1 \times I_2 \times \dots \times I_K}$, where $I_{k}$ denotes the dimension in mode-$k$. The entry in the $i_k$th index in mode-$k$ for $k=1,\dots, N$ is denoted as $\mathcal{X}_{i_1,i_2,\dots,i_K}$.

\subsection{Multilinear Algebra Concepts}
\begin{defn}[Mode-$k$ Fiber and Mode-$k$ Unfolding]\label{def1}
	The mode-$k$ fiber of a tensor $\mathcal{X} \in \mathbb{R}^{I_1 \times I_2 \times \dots \times I_K}$ is a vector of $I_k$-dimensional, given by fixing every index but $i_k$. The mode-$k$ unfolding of $\mathcal{X}$, also known as mode-$k$ matricization, transforms the tensor $\mathcal{X}$ to matrix $\mathbf{X}_{(k)}$, which is formed by arranging the mode-$k$ fibers as columns. The shape of $\mathbf{X}_{(k)}$ is $\mathbb{R}^{I_k \times I_{\bar{k}}}$ with $I_{\bar{k}}=\prod_{i=1,i \neq k}^{K} I_i$.
\end{defn}

\begin{defn}[Mode-$k$ Product]\label{def2}
	The mode-$k$ product between a tensor $\mathcal{X}=[x_{i_1},\dots , x_{i_K}] \in  \mathbb{R}^{I_1 \times \dots I_K}$ and a matrix $\mathbf{W}\in \mathbb{R}^{J_{k}\times I_k}$ is another tensor of size $I_1\times \dots \times J_{k}\times \dots \times I_K$ and denoted by $\mathcal{X} \times_{k} \mathbf{W}$. The element of $\mathcal{X} \times_{k} \mathbf{W}$ is defined as $[\mathcal{X}\times_{k}\mathbf{W}]_{i_1, \dots , i_{k-1}, j_k, i_{k+1},\dots, i_K}=\sum_{i_k=1}^{I_K}[\mathcal{X}]_{i_1,\dots,i_{k-1},i_k,\dots, i_K}[\mathbf{W}]_{j_k,i_k}$.
\end{defn}

For convenience, we denote $\mathcal{X}\times_1\mathbf{W}_1\times\dots\times_K
\mathbf{W}_K$ by $\mathcal{X} \prod_{k=1}^{K}\times_k\mathbf{W}_k$.

One of the nice properties of mode-$k$ product is that the result of the projection does not depend on the order of projection, i.e.
\begin{equation}\label{eq2}
\big(\mathcal{X}\times_{k_1}\mathbf{A}\big)\times_{k_2}\mathbf{B}=\big(\mathcal{X}\times_{k_2}\mathbf{B}\big)\times_{k_1}\mathbf{A}.
\end{equation}

The above property allows efficient computation of the projection by selecting the order of computation.

\subsection{Multilinear filter as generic feature extractor}
\begin{figure*}[tp]\label{figure1}
  \centering
  \includegraphics[width=\textwidth]{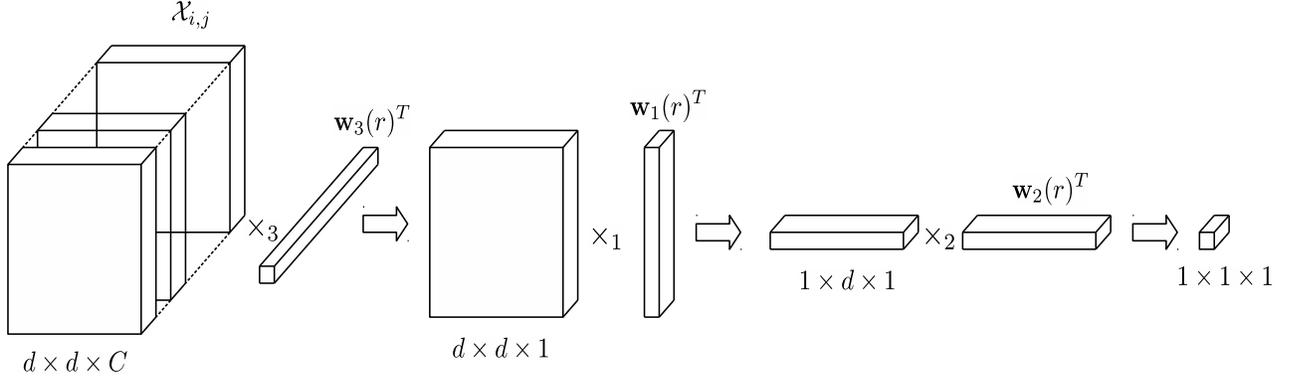}
  \caption{Illustration of the proposed multilinear mapping according to equation (\ref{eq6}) in sequence: mode-$3$, mode-$1$ and mode-$2$}
\end{figure*}

Let $\mathcal{X}_{i,j}\in \mathbb{R}^{d\times d\times C}$ and $\mathcal{W}\in \mathbb{R}^{d\times d\times C}$ denote the input patch centered at spatial location $(i,j)$ and the convolution kernel respectively. At the core of a classic CNN, each convolution kernel operates as a feature extractor sliding through the input tensor to generate a higher level representation. Specifically, the kernel performs the following linear mapping:
\begin{equation}\label{eq3}
\mathcal{Y}_{i,j}=f\big(\mathcal{X}_{i,j};\mathcal{W};b\big)= \braket{\mathcal{X}_{i,j} ,\mathcal{W}} + b,
\end{equation}
where $\mathcal{Y}_{i,j}$ and $b$ denotes the response at $(i,j)$ and the intercept term respectively. $\braket{\boldsymbol{\cdot},\boldsymbol{\cdot}}$ denotes the dot-product between two tensors. After the above linear projection, a nonlinearity is applied to $\mathcal{Y}_{i,j}$ using the layer's activation function.

We propose to replace the above linear projection by the following multilinear mapping:
\begin{equation}\label{eq4}
\mathcal{Y}_{i,j}=\tilde{f}\big(\mathcal{X}_{i,j};\tilde{W};b\big)=\sum_{r=1}^{R}\mathcal{X}_{i,j}\prod_{k=1}^{3} \times_k \mathbf{w}_{k}(r)^{T} + b,
\end{equation}
where $R$ is the rank hyper-parameter of the projection and $\mathbf{w}_{k}(r)$ denotes the projection along mode-$k$. In our case, $\mathbf{w}_{1}(r) \in \mathbb{R}^{d},\mathbf{w}_{2}(r) \in \mathbb{R}^{d}, \mathbf{w}_{3}(r) \in \mathbb{R}^{C}, \forall r=1,\dots,R$.

Since the mapping in Eq. (\ref{eq4}) operates on similar input patch and yields a scalar response as a linear mapping does in CNNs, the proposed multilinear mapping acts as a generic feature extractor and can be incorporated into any design of the CNN topology, such as AlexNet \cite{krizhevsky2012imagenet}, VGG \cite{simonyan2014very}, Inception \cite{szegedy2017inception} or ResNet \cite{he2016deep}. In addition, since the mapping in Eq. (\ref{eq4}) is differentiable with respect to each individual weight vector $\mathbf{w}_{k}(r)$, the resulting network architecture can be trained in an end-to-end fashion by back propagation algorithm. We hereby denote the layer employing our proposed multilinear mapping as MLconv.

Recently, mode-$k$ multiplication has been introduced as a tensor contraction layer in \cite{kossaifi2017tensor} to project the entire input layer as a high-order tensor to another tensor. This is fundamentally different from our approach since the tensor contraction layer is a global mapping which does not incorporate sparse connectivity and parameter sharing principles. In general, mode-$k$ multiplication can be applied to an input patch/volume to output another tensor instead of a scalar as in our proposal. We restrict the multilinear projection in the form of Eq. (\ref{eq4}) to avoid the increase in the output dimension which leads to computation overhead in the next layer. Moreover, tensor unfolding operation required to perform the multilinear projection that transforms a tensor to another tensor will potentially increase the computation. On the contrary, our proposed mapping is a special case of the general multilinear mapping using mode-$k$ product in which the output tensor degenerates to a scalar. This special case allows efficient computation of the projection, as shown in the next section.

\subsection{Memory and Computation Complexity}\label{SS:MemoryComputationComplexity}
One the most obvious advantages of the mapping in Eq. (\ref{eq4}) is that it requires far fewer parameters to estimate the model, compared to the linear mapping in a CNN. In a CNN utilizing the mapping in Eq. (\ref{eq3}), a layer with $N$ kernels requires the storage of $d^2CN$ parameters. On the other hand, a similar layer configuration with $N$ mappings utilizing the projection in Eq. (\ref{eq4}) requires only $R(2d+C)N$ parameters. The gain ratio is:
\begin{equation}\label{eq5}
\frac{d^2C}{R(2d+C)}.
\end{equation}

As compared to a similar CNN topology, the memory reduction utilizing the mapping in Eq. (\ref{eq4}) varies for different layers. The case where $C >> d$ (which is the usual case) leads to a gain ratio approximately equal to $d^2/R$. In our experiments, we have seen that with $d=3$ and $R=2$ in all layers, memory reduction is approximately $4\times$, while having competitive performance compared to a CNN with similar network topology.

Let us denote by $\mathcal{X}_l \in \mathbb{R}^{X\times Y\times C}$ and $\mathcal{W}_l \in \mathbb{R}^{d\times d\times C\times N}$ the input and $N$ kernels of the $l$-th convolutional layer having $C$ input feature maps and  $N$ output feature maps. In addition, we assume zero-padding and sliding window with stride of $1$. By using linear projection as in case of CNN, the computational complexity of this layer is $O(d^2XYCN)$. Before evaluating the computational cost of a layer using the proposed method, it should be noted that the projection in Eq. (\ref{eq4}) can be efficiently computed by applying three consecutive convolution operations. Details of the convolution operations depend on the order of three modes. Therefore, although the result of the mapping in Eq. (\ref{eq4}) is independent of the order of mode-$k$ projection, the computational cost actually depends on the order of projections. For $C >> d$, it is computationally more efficient to first perform the projection in mode-$3$ in order to reduce the number of input feature maps for subsequent mode-$1$ and mode-$2$ projection:
\begin{equation}\label{eq6}
\mathcal{Y}_{i,j}=\sum_{r=1}^{R}\mathcal{X}_{i,j}\times_3 \mathbf{w}_{3}(r)^{T}\times_1 \mathbf{w}_{1}(r)^{T}\times_2 \mathbf{w}_{2}(r)^{T} + b.
\end{equation}

The response $\mathcal{Y}_{i,j}$ in Eq. (\ref{eq6}) is the summation of $R$ independent projections with each projection corresponding to the following three consecutive steps, as illustrated in Figure \ref{figure1}:
\begin{itemize}
	\item Projection of $\mathcal{X}_{i,j}$ along the third mode which is the linear combination of $C$ input feature maps. The result is a tensor $\mathcal{X}_{i,j}^{(3)}$ of size $d\times d\times 1$.
	\item Projection of $\mathcal{X}_{i,j}^{(3)}$ along the first mode which is the linear combination of $d$ rows. The result is a tensor $\mathcal{X}_{i,j}^{(1)}$ of size $1\times d\times 1$.
	\item Projection of $\mathcal{X}_{i,j}^{(1)}$ along the second mode which is the linear combination of $d$ elements.
\end{itemize}
With the aforementioned configuration of the $l$-th layer, the computational complexity of the $l$-th MLconv layer utilizing our multilinear mapping is as follows:
\begin{itemize}
	\item Mode-$3$ projection that corresponds to applying $NR$ convolutions to the input $\mathcal{X}_l$ with kernels of size $1\times 1\times C$ elements, having computational complexity of $O(XYCNR)$. The output of the projection along the third mode is a tensor of size $X\times Y\times NR$.
	\item Mode-$1$ projection is equivalent to applying convolution with one $d\times 1\times NR$ separable convolution kernel, having complexity of $O(dXYNR)$. This results in a tensor of size $X\times Y\times NR$.
	\item Mode-$2$, similar to mode-$1$ projection, can be computed by applying convolution with one $1\times d\times NR$ separable convolution kernel, requiring $O(dXYNR)$ computation. This results in a tensor of size $X\times Y\times NR$. By summing over $R$ ranks, we arrive at the output of layer $l$ of size $X\times Y\times N$.
\end{itemize}

The total complexity of layer $l$ using our proposed mapping is thus $O(XYNR(C+2d))$. Compared to linear mapping, our method achieves computational gain of:
\begin{equation}\label{eq7}
\frac{d^2C}{R(C+2d)}.
\end{equation}

From Eqs. (\ref{eq5}) and (\ref{eq7}), we can conclude that the proposed feature extractor achieves approximately $d^2/R$ savings in both computation and memory when $C>>d$.

\subsection{Initialization with pre-trained CNN}\label{SS:Initializationpre-trained}
The proposed mapping in Eq. (\ref{eq4}) can be viewed as a constrained form of convolution kernel as follows:
\begin{equation}\label{eq8}
\mathcal{Y}_{i,j}=\braket{\mathcal{X}_{i,j}, \tilde{\mathcal{W}}}+b,
\end{equation}
where $\tilde{\mathcal{W}}$ is expressed in Kruskal form $\tilde{\mathcal{W}}=\sum_{r=1}^{R}\mathbf{w}_1(r)\circ \mathbf{w}_2(r) \circ \mathbf{w}_3(r)$ as the outer-product of the corresponding projection in three modes. By calculating $\mathcal{Y}_{i,j}$ using mode-$k$ product definition as in Eq. (\ref{eq4}) and using dot-product as in Eq. (\ref{eq8}), the equivalance of Eq. (\ref{eq4}) and Eq. (\ref{eq8}) can be found \cite{kolda2009tensor}. 

Consequently, a convolution layer can be converted to an MLconv layer by decomposing each $3D$ convolution filter into Kruskal form using any CP decomposition method \cite{kolda2009tensor}. It should be noted here that, since there is no closed-form solution of the CP decomposition, such a conversion corresponds to an approximation step. Under this perspective, a pre-trained CNN can be used to initialize our network structure to speed up the training process. However, as we will show in the experimental section, random initialization of multilinear filters can lead to better performance.

In addition to an initialization scheme, Eq. (\ref{eq8}) also complements our proposed mapping with an efficient computation strategy when $R$ is large. The computation cost discussed in the previous subsection depends linearly with parameter $R$. When $R$ is large, it is more efficient to compute the mapping according to Eq. (\ref{eq8}) by first calculating $\tilde{\mathcal{W}}$ and then convolving the input with $\tilde{\mathcal{W}}$. The computational complexity of the first step is $O(d^2CRN)$ while for the convolution step is $O(d^2XYCN)$, resulting to an overall complexity of $O(d^2CRN+d^2XYCN)$ for the entire layer. The ratio between normal convolution layer and MLconv layer using this computation strategy is:
\begin{equation}\label{eq9}
\frac{XY}{R+XY}.
\end{equation}
It is clear that $XY$ is usually much larger than $R$, therefore, the increase in computation as compared to normal convolution is marginal. Following this calculation strategy, a rank $6$ network is marginally slower than a rank $1$ network or a CNN. This will be demonstrated in our experiment section. In conclusion, the computation method discussed in this subsection allows the scalability of our proposed mapping when $R$ is large while previous subsection proposes an efficient computation scheme that allows computation savings when $R$ is small. Overall, we can conclude that the computation of the proposed layer structure is efficient while, as will be shown in the experimental evaluation, changing the rank of the adopted tensor definitions can increase performance.

\section{Experiments}
In this section, we provide experimental results to support the theoretical analysis in section \ref{S:ProposedMethod}. The experimental protocol and datasets are described first, followed by the discussion of the experimental results.

\subsection{Network Topology}
Traditional CNN topology consists of two modules: feature extractor module and classifier module. Several convolution and pooling layers stacked on top of each other act as feature extractor while one or two fully-connected layers act as the classifier. In order to evaluate the effectiveness of the proposed multilinear filter, we constructed the network architecture with only feature extractor layers, i.e. convolution layer or MLconv layer together with pooling layer while skipping fully-connected layer. As the name suggests, fully-connected layer has dense connections, accounting for large number of parameters in the network while being prone to overfitting. Moreover, a powerful and effective feature extractor module is expected to produce a highly discriminative latent space in which the classification task is made simple. Such fully-convolutional networks have attracted much attention lately due to their compactness and excellent performance in image-related problems like semantic segmentation, object localization and classification \cite{lin2013network, redmon2016you, ronneberger2015u}

The configuration of the baseline network adopted in our experiment benchmark is shown in Table \ref{t1} where $3\times3\times N$ denotes $N$ kernels with $3\times 3$ spatial dimension, BN denotes Batch Normalization \cite{ioffe2015batch} and LReLU denotes Leaky Rectified Linear Unit \cite{maas2013rectifier} with $\alpha=0.2$. Our baseline architecture is similar to the one proposed in \cite{springenberg2014striving}  with four key differences. Firstly, we choose to retain a proper pooling layer instead of performing convolution with a stride of $2$ as proposed in \cite{springenberg2014striving}. Secondly, Batch Normalization was applied after every convolution layer except the last one where the output goes through softmax to produce the class probability. In addition, LReLU activation unit was applied to the output of batch normalization. It has been shown that the adoption of BN and LReLU speeds up the learning process of the network by being more tolerant to the learning rate with the possibility of arriving at better minimas \cite{ioffe2015batch,xu2015empirical}.
\begin{table}[t]
	\begin{center}
		\caption{Baseline Network Architecture}\label{t1}
		\resizebox{0.6\linewidth}{!}{
			\begin{tabular}{|c|}
				\hline
				Input layer 	\\ \hline
				$3\times3\times96$ - BN - LReLU \\ \hline
				$3\times3\times96$ - BN - LReLU \\ \hline
				$3\times3\times96$ - BN - LReLU \\ \hline
				$2\times2$ MaxPooling \\ \hline		
				$3\times3\times192$ - BN - LReLU \\ \hline
				$3\times3\times192$ - BN - LReLU \\ \hline
				$3\times3\times192$ - BN - LReLU \\ \hline
				$2\times2$ MaxPooling \\ \hline
				$3\times3\times192$ - BN - LReLU \\ \hline
				$1\times1\times192$ - BN - LReLU \\ \hline
				$1\times1\times \#class$  LReLU \\ \hline
				Global Average over spatial dimension \\ \hline
				softmax activation \\ \hline		
				
			\end{tabular}
		}
	\end{center}
\end{table}

Based on the configuration of the network topology, we compare the performance between standard linear convolution kernel (CNN), our proposed multilinear kernel (MLconv) and the low-rank (LR) structure proposed in \cite{tai2015convolutional}. The last two $1\times1$ convolution layers were not replaced by LR or MLconv layer. It should be noted that BN and LReLU are applied to all three competing structures in our experiments while in \cite{tai2015convolutional}, BN was not applied to the baseline CNN which could potentially lead to biased result.

\subsection{Datasets}
\subsubsection{CIFAR-10 and CIFAR-100}
CIFAR dataset \cite{krizhevsky2009learning} is an object classification dataset which consists of $50000$ color images for training and $10000$ for testing with the resolution $32\times32$ pixels. CIFAR-10 refers to the 10-class classification problem of the dataset in which each class has $5000$ images for training and $1000$ images for testing while CIFAR-100 refers to a more fine-grained classification of the images into $100$ classes.

\subsubsection{SVHN}
SVHN \cite{netzer2011reading} is a well-known dataset for hand-written digit recognition problem which consists of more than $600k$ images of house numbers extracted from natural scenes with varying number of samples from each class. This dataset poses a much harder character recognition problem as compared to the MNIST dataset \cite{lecun1998gradient}. We used $32\times32$ cropped images provided by the database from which each individual image might contain some distracting digits on the sides.

\begin{table*}[bp]	
	\begin{center}
		\caption{CIFAR-10 Classification error (\%)}\label{t2}
		\resizebox{1.0\textwidth}{!}{
			\begin{tabular}{|l|c|c|c|c|c|c|c|c|}\cline{2-9}
				\multicolumn{1}{c|}{}
				& CNN 	& MLconv1	& LR26	& MLconv2	& LR53	& MLconv4	& LR106	& MLconv6 \\ \hline
				Scratch		& $7.47$		& $8.54$	& $9.14$	& $7.68$	& $8.31$	& $7.34$	& $8.00$	& $7.30$  \\ \hline
				Pretrained	& $-$				& $8.17$	& $8.64$	& $7.76$	& $7.49$	& $7.38$	& $7.10$	& $7.06$  \\ \hline
				\# Parameters & $\approx 1.38M$ & $\approx 0.20M$ & $\approx 0.20M$ & $\approx 0.35M$ & $\approx 0.35M$ & $\approx 0.65M$ & $\approx 0.64M$ & $\approx 0.97M$ \\ \hline	
				
			\end{tabular}
	}	
	\end{center}
\end{table*}

\begin{table*}[bp]	
	\begin{center}
		\caption{CIFAR-100 Classification error (\%)}\label{t3}
		\resizebox{1.0\textwidth}{!}{
			\begin{tabular}{|l|c|c|c|c|c|c|c|c|}\cline{2-9}
				\multicolumn{1}{c|}{}
				& CNN  	& MLconv1	& LR26		& MLconv2	& LR53		& MLconv4	& LR106		& MLconv6 \\ \hline
				Scratch		& $29.60$		& $31.32$	& $35.79$	& $29.10$	& $31.45$	& $28.27$	& $30.11$	& $28.08$  \\ \hline
				Pretrained	& $-$				& $31.88$	& $33.98$	& $29.86$	& $30.00$	& $28.45$	& $28.40$	& $28.51$  \\ \hline
				\# Parameters & $\approx 1.39M$ & $\approx 0.21M$ & $\approx 0.21M$ & $\approx 0.37M$ & $\approx 0.37M$ & $\approx 0.67M$ & $\approx 0.67M$ & $\approx 0.98M$ \\ \hline	
				
			\end{tabular}
		}	
	\end{center}
\end{table*}

\subsection{Experimental settings}
All networks were trained using both SGD optimizer \cite{rumelhart1988learning} as well as Adam \cite{kingma2014adam}. While the proposed structure tends to arrive at better minimas with Adam, this is not the case for the other two methods. For SGD optimizer, the momentum was fixed to $0.9$. We adopted two sets of learning rate schedule $SC_{1}=\{0.01,0.005,0.001,0.0005,0.0001\}$ and $SC_2=\{0.01,0.001,0.0001\}$. Each schedule has initial learning rate $\gamma=0.01$ and decreases to the next value after $E$ epochs where $E$ was cross-validated from the set $\{40,50,60,100,120\}$. We trained each network with maximum of $300$ and $100$ epochs for CIFAR and SVHN respectively. The batch size was fixed to $200$ samples for all competing networks.

Regarding data augmentation, for CIFAR dataset, random horizontally flipped samples were added as well as random translation of the images by maximum $5$ pixels were performed during the training process; for SVHN dataset, only random translation of maximum $5$ pixels was performed. For both dataset, no further preprocessing step was applied.

Regarding regularization, both weight decay and max-norm \cite{srivastava2014dropout} are individually and together exploited in our experiments. Max-norm regularizer was introduced in \cite{srivastava2014dropout} where it was used together with Dropout. During the training process, the $l_2$ norm of each individual filter is constrained to lie inside the ball of a given radius $r$ which was cross-validated from the set $\{1.0,2.0,4.0,6.0,8.0\}$. The weight decay hyper-parameter $\lambda$ was searched from the set $\{0.001,0.0005\}$. In addition, Dropout with $p_i=0.2$ was applied to the input and Dropout with $p = p_o$  was applied to the output of all pooling layers with the optimal $p_o$ obtained from the set $\{0.1,0.2,0.3,0.4,0.5\}$. Due to the differences between the three competing structures, we observed that while the baseline CNN and LR networks work well with weight decay, applying weight decay to the proposed network structure tends to drive all the weight values close to zeros when $\lambda$ is large, or the regularization effect is marginal when using a small value for $\lambda$, leading to the exhaustive search of suitable hyper-parameter $\lambda$. On the other hand, max-norm regularization works well with our method without being too sensitive to the performance.

For MLconv and LR structures, we experimented with several values for the rank parameter, namely $R$ for the proposed mapping and $K$ in Eq. (\ref{eq1}) from \cite{tai2015convolutional}. In all of our experiments, we made no attempt to optimize $R$ and $K$ for each individual filter and layer in order to get the maximal compact structure, since such an approach is impractical in real cases. We instead used the same rank value throughout all layers. The experiments are, hence, different from \cite{tai2015convolutional} where the authors reported performance for different values of $K$ at each layer without discussing the rank selection method. The experiments were conducted with $R={1,2,4,6}$ and the corresponding structures are denoted as MLconv1, MLconv2, MLconv4, MLconv6. The values of $K$ are selected so that the number of parameters in an LR network is similar to the number of parameters of its MLconv counterpart with given $R$. The corresponding LR structures are denoted as LR26, LR53 and LR106, where the number denotes the value of $K$. We did not perform experiments with $K=159$, which corresponds to $R=6$, since training the network is computationally much slower and falls out of the objective of this paper.

All of three competing structures training from scratch were initialized with random initialization scheme proposed in \cite{he2015delving}. We additionally trained MLconv and LR structure with weights initialized from an optimal pre-trained CNN on CIFAR dataset. The aforementioned protocols were also applied for this configuration. The weights of MLconv were initialized with CP decomposition using canonical alternating least square method \cite{kolda2009tensor}, while for the LR structure we followed the calculation proposed in \cite{tai2015convolutional}.

\begin{figure}[t!]
	\centering
	\includegraphics[width=0.5\textwidth]{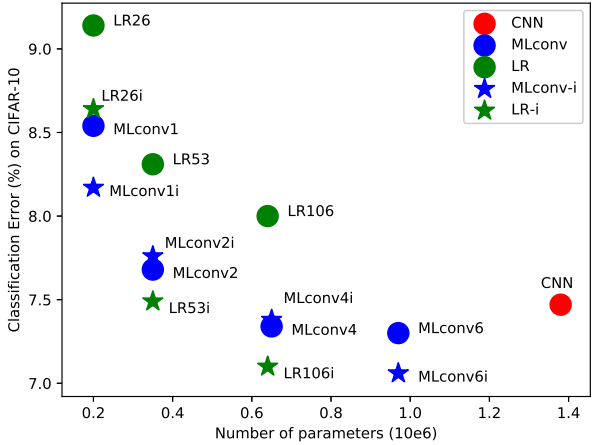}
	\caption{Model size versus Classification Error on CIFAR-10 for different structures. MLconv and LR network initialized with CNN marked with "i" at last }\label{figure2}
\end{figure}

\subsection{Experimental results}

\begin{figure}[t!]
	\centering
	\includegraphics[width=0.5\textwidth]{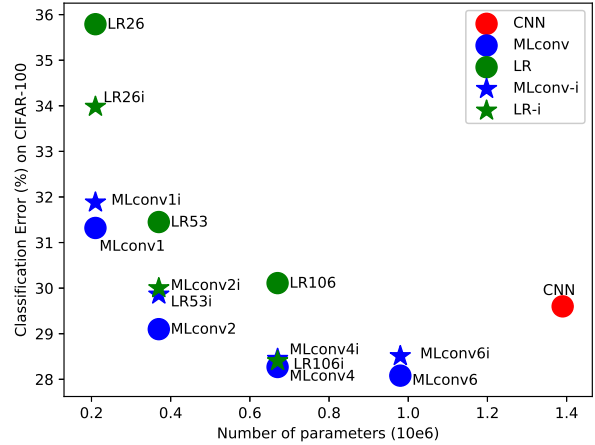}
	\caption{Model size versus Classification Error on CIFAR-100 for different structures. MLconv and LR network initialized with CNN marked with "i" at last}\label{figure3}
\end{figure}

After obtaining the optimal hyper-parameter values, each network was trained for five times and the median value is reported. The second row of Tables \ref{t2} and \ref{t3} shows the classification errors of all competing methods trained from scratch on CIFAR-10 and CIFAR-100, respectively, while the third row shows the performance when initialized with a pre-trained CNN. The last row reports the model size of each network. As can be seen from both Tables \ref{t2} and \ref{t3}, using the proposed multi-linear filters leads to a $2\times$ reduction in memory, while outperforming the standard convolution filters in both coarse and fine-grained classification in CIFAR datasets. More interestingly, in CIFAR-100, a rank $4$ multi-linear filter network attains an improvement over $1\%$. As we increase the number of projections in each mode to $6$, i.e. when using $R=6$, the performance of the network increases by a small margin. In both CIFAR-10 and CIFAR-100, constraining $R=2$ gains $4\times$ memory reduction while keeping the performance relatively closed to the baseline CNN with less than $0.5\%$ increment in classification error. Further limiting $R$ to $1$ maximizes the parameter reduction to nearly $7\times$ with the cost of $1.07\%$ and $1.72\%$ increase in error rate for CIFAR-10 and CIFAR-100, respectively. A graphical illustration of the compromise between number of network's parameters and classification error on CIFAR-10 and CIFAR-100 is illustrated in Figures \ref{figure2} and \ref{figure3}, respectively.

The classification error of each competing network trained from scratch on SVHN dataset is shown in Table \ref{t5}. Using our proposed MLconv layers, we achieved $4\times$ reduction in model size while slightly outperforming CNN. At the most compact configuration of MLconv structure, i.e. MLconv1, we only observed a small increment of $0.12\%$ in classification error as compared to CNN baseline. As we increased the complexity of MLconv layers, little improvement was seen with MLconv4 while MLconv6 layers became slightly overfitted.

Comparing the proposed multi-linear filter with the low rank structure LR, all configurations of MLconv network significantly outperform their LR counterparts. Specifically, in the most compact configuration, MLconv1 is better than LR26 by $0.6\%$ and $4.47\%$ on CIFAR-10 and CIFAR-100, respectively. The margin shrinks as the complexity increases but the proposed structure consistently outperforms LR when training the network from scratch. Similar comparison results can be observed on SVHN dataset: using MLconv layers obtained lower classification errors as compared to LR layers at all complexity configurations. As opposed to the experimental results reported in \cite{tai2015convolutional}, we observed inferior results of the LR structure compared to standard CNN when training from scratch. The difference might be attributed to two main reasons: we incorporated batch normalization into the baseline CNN which could potentially improve the performance of the baseline CNN; our baseline configuration has no fully-connected layer to solely benchmark the efficiency of different filter structures as a feature extractor. 

\begin{table}[t!]
	\begin{center}
		\caption{SVHN Classification error}\label{t5}
		\resizebox{0.7\linewidth}{!}{
			\begin{tabular}{|l|c|c|}\cline{2-3}
				\multicolumn{1}{c|}{}
				& Error (\%) 			& \#Parameters \\ \hline
				CNN		& $1.80$			& $\approx 1.38M$	\\ \hline \hline	
				MLconv1	& $1.92$ 			& $\approx 0.20M$ 	\\ \hline
				LR26	& $1.96$			& $\approx 0.20M$	\\ \hline \hline
				
				MLconv2	& $1.76$			& $\approx 0.35M$ \\ \hline
				LR53	& $1.85$			& $\approx 0.35M$		\\ \hline \hline
				
				MLconv4	& $1.75$			& $\approx 0.65M$ \\ \hline
				LR106	& $1.78$			& $\approx 0.64M$	\\ \hline \hline
				
				MLconv6	& $1.80$			& $\approx 0.97M$	\\ \hline	
								
			\end{tabular}
		}
	\end{center}
\end{table}

\begin{table*}[bp]	
	\begin{center}
		\caption{Forward Computation Time on CIFAR10 (normalized with respect to Conv)}\label{t4}
		\resizebox{1.0\textwidth}{!}{
			\begin{tabular}{|l|c|c|c|c|c|c|c|c|c|c|c|c|}\cline{2-12}
				\multicolumn{1}{c|}{}
				& MLconv1 	& LR26		& MLconv2	& LR53	& MLconv4	& LR106	 & MLconv6	& MLconv1*	& MLconv2*	& MLconv4*	& MLconv6* \\ \hline
				Theory			& $8.14\times$	& $6.48\times$	& $4.11\times$	& $3.20\times$	& $2.06\times$	& $1.60\times$	& $1.38\times$ 
				& $0.997\times$ & $0.993\times$ & $0.987\times$ & $0.980\times$ \\ \hline
				Implementation	& $1.77\times$	& $1.79\times$	& $1.33\times$	& $1.54\times$	& $1.03\times$	& $1.23\times$	& $0.74\times$  
				& $0.990\times$ & $0.980\times$ & $0.960\times$ & $0.930\times$  \\ \hline	
				
			\end{tabular}
		}	
	\end{center}
\end{table*}

One interesting phenomenon was observed when we initialized MLconv and LR with a pre-trained CNN. For the LR structure, most configurations enjoy substantial improvement by initializing the network with weights decomposed from a pre-trained CNN on CIFAR dataset. The contrary happens for our proposed MLconv structure, since most configurations observe a degradation in performance. This can be explained by the fact that LR structure was designed to approximate each individual 2D convolution filter at every input feature map and the resulting structure comes with a closed-form solution for the approximation. With good initialization from a CNN, the network easily arrived at a good minimum while training a low-rank setting from scratch might have difficulty at achieving a good local minimum. Although the proposed mapping can be viewed as a form of convolution filter, the mapping in Eq. (\ref{eq4}) embeds a multi-linear structure, hence possessing certain degree of difference. Initializing the proposed mapping by applying CP decomposition, which has no closed-form solution, may lead the network to a suboptimal state.

Table \ref{t4} reports the average forward propagation time of a single sample measured on CPU for all three network structures on CIFAR-10. The second and third columns report the theoretical and actual speed-ups, respectively, measured by the number of multiply-accumulate operations normalized with respect to their convolution counterparts. For the proposed MLconv structure, we report the computation cost of both calculation strategies discussed in Section \ref{S:ProposedMethod}. We refer to the first calculation strategy using the separable convolution as Scheme1, while the latter one using normal convolution as Scheme2. Results from Scheme2 are denoted with the asterisk. All the networks are implemented using Keras library \cite{chollet2015} with Tensorflow \cite{tensorflow2015paper} backend. It is clear that there is a gap between theoretical speed-up and actual speed-up, especially for the proposed structure implemented by an unoptimized separable convolution operation. In fact, at the time of writing, implementation of separable convolution operation is still missing in most libraries, not to mention efficient implementation. On the contrary, results from Scheme2 using normal convolution show a near perfect match between theory and implementation. This is due to the fact that normal convolution operation has been efficiently implemented and optimized in most popular libraries. This also explains why the computation gain of LR structure is inferior to MLconv structure (Scheme1) in theory but similar to ours in practice since LR structure is realized by normal convolution operation. The last four columns of Table \ref{t4} additionally prove the scalability of Scheme2 with respect to the hyper-parameter rank $R$ as discussed in Section \ref{SS:Initializationpre-trained}.

\section{Conclusions}
In this paper, we proposed a multilinear mapping to replace the conventional convolution filter in Convolutional Neural Networks. The resulting structure's complexity can be flexibly controlled by adjusting the number of projections in each mode through a hyper-parameter $R$. The proposed mapping comes with two computation schemes which either allow memory and computation reduction when $R$ is small, or the scalability when $R$ is large. Numerical results showed that with far fewer parameters, architectures employing our mapping could outperform standard CNNs. This are promising results and opens future research directions focusing on optimizing parameter $R$ on individual convolution layers to achieve the most compact structure and performance.

\bibliography{template}

\begin{thebibliography}{10}

\bibitem{girshick2014rich}
R.~Girshick, J.~Donahue, T.~Darrell, and J.~Malik, ``Rich feature hierarchies
  for accurate object detection and semantic segmentation,'' in {\em
  Proceedings of the IEEE conference on computer vision and pattern
  recognition}, pp.~580--587, 2014.

\bibitem{redmon2016you}
J.~Redmon, S.~Divvala, R.~Girshick, and A.~Farhadi, ``You only look once:
  Unified, real-time object detection,'' in {\em Proceedings of the IEEE
  Conference on Computer Vision and Pattern Recognition}, pp.~779--788, 2016.

\bibitem{waris2017cnn}
M.~A. Waris, A.~Iosifidis, and M.~Gabbouj, ``Cnn-based edge filtering for
  object proposals,'' {\em Neurocomputing}, 2017.

\bibitem{hinton2012deep}
G.~Hinton, L.~Deng, D.~Yu, G.~E. Dahl, A.-r. Mohamed, N.~Jaitly, A.~Senior,
  V.~Vanhoucke, P.~Nguyen, T.~N. Sainath, {\em et~al.}, ``Deep neural networks
  for acoustic modeling in speech recognition: The shared views of four
  research groups,'' {\em IEEE Signal Processing Magazine}, vol.~29, no.~6,
  pp.~82--97, 2012.

\bibitem{graves2013speech}
A.~Graves, A.-r. Mohamed, and G.~Hinton, ``Speech recognition with deep
  recurrent neural networks,'' in {\em Acoustics, speech and signal processing
  (icassp), 2013 ieee international conference on}, pp.~6645--6649, IEEE, 2013.

\bibitem{zabihi2016heart}
M.~Zabihi, A.~B. Rad, S.~Kiranyaz, M.~Gabbouj, and A.~K. Katsaggelos, ``Heart
  sound anomaly and quality detection using ensemble of neural networks without
  segmentation,'' in {\em Computing in Cardiology Conference (CinC), 2016},
  pp.~613--616, IEEE, 2016.

\bibitem{an2014deep}
X.~An, D.~Kuang, X.~Guo, Y.~Zhao, and L.~He, ``A deep learning method for
  classification of eeg data based on motor imagery,'' in {\em International
  Conference on Intelligent Computing}, pp.~203--210, Springer, 2014.

\bibitem{tsantekidis2017using}
A.~Tsantekidis, N.~Passalis, A.~Tefas, J.~Kanniainen, M.~Gabbouj, and
  A.~Iosifidis, ``Using deep learning to detect price change indications in
  financial markets,'' in {\em European Signal Processing Conference (EUSIPCO),
  Kos, Greece}, 2017.

\bibitem{tsantekidis2017forecasting}
A.~Tsantekidis, N.~Passalis, A.~Tefas, J.~Kanniainen, M.~Gabbouj, and
  A.~Iosifidis, ``Forecasting stock prices from the limit order book using
  convolutional neural networks,'' in {\em Business Informatics (CBI), 2017
  IEEE 19th Conference on}, vol.~1, pp.~7--12, IEEE, 2017.

\bibitem{lecun1998gradient}
Y.~LeCun, L.~Bottou, Y.~Bengio, and P.~Haffner, ``Gradient-based learning
  applied to document recognition,'' {\em Proceedings of the IEEE}, vol.~86,
  no.~11, pp.~2278--2324, 1998.

\bibitem{he2016deep}
K.~He, X.~Zhang, S.~Ren, and J.~Sun, ``Deep residual learning for image
  recognition,'' in {\em Proceedings of the IEEE conference on computer vision
  and pattern recognition}, pp.~770--778, 2016.

\bibitem{szegedy2015going}
C.~Szegedy, W.~Liu, Y.~Jia, P.~Sermanet, S.~Reed, D.~Anguelov, D.~Erhan,
  V.~Vanhoucke, and A.~Rabinovich, ``Going deeper with convolutions,'' in {\em
  Proceedings of the IEEE conference on computer vision and pattern
  recognition}, pp.~1--9, 2015.

\bibitem{han2015deep}
S.~Han, H.~Mao, and W.~J. Dally, ``Deep compression: Compressing deep neural
  networks with pruning, trained quantization and huffman coding,'' {\em arXiv
  preprint arXiv:1510.00149}, 2015.

\bibitem{guo2016dynamic}
Y.~Guo, A.~Yao, and Y.~Chen, ``Dynamic network surgery for efficient dnns,'' in
  {\em Advances In Neural Information Processing Systems}, pp.~1379--1387,
  2016.

\bibitem{chen2015compressing}
W.~Chen, J.~T. Wilson, S.~Tyree, K.~Q. Weinberger, and Y.~Chen, ``Compressing
  convolutional neural networks,'' {\em arXiv preprint arXiv:1506.04449}, 2015.

\bibitem{wen2016learning}
W.~Wen, C.~Wu, Y.~Wang, Y.~Chen, and H.~Li, ``Learning structured sparsity in
  deep neural networks,'' in {\em Advances in Neural Information Processing
  Systems}, pp.~2074--2082, 2016.

\bibitem{gong2014compressing}
Y.~Gong, L.~Liu, M.~Yang, and L.~Bourdev, ``Compressing deep convolutional
  networks using vector quantization,'' {\em arXiv preprint arXiv:1412.6115},
  2014.

\bibitem{lin2016fixed}
D.~Lin, S.~Talathi, and S.~Annapureddy, ``Fixed point quantization of deep
  convolutional networks,'' in {\em International Conference on Machine
  Learning}, pp.~2849--2858, 2016.

\bibitem{tai2015convolutional}
C.~Tai, T.~Xiao, Y.~Zhang, X.~Wang, and E.~Weinan, ``Convolutional neural
  networks with low-rank regularization,'' {\em arXiv preprint
  arXiv:1511.06067}, 2015.

\bibitem{denton2014exploiting}
E.~L. Denton, W.~Zaremba, J.~Bruna, Y.~LeCun, and R.~Fergus, ``Exploiting
  linear structure within convolutional networks for efficient evaluation,'' in
  {\em Advances in Neural Information Processing Systems}, pp.~1269--1277,
  2014.

\bibitem{jaderberg2014speeding}
M.~Jaderberg, A.~Vedaldi, and A.~Zisserman, ``Speeding up convolutional neural
  networks with low rank expansions,'' {\em arXiv preprint arXiv:1405.3866},
  2014.

\bibitem{ioannou2015training}
Y.~Ioannou, D.~Robertson, J.~Shotton, R.~Cipolla, and A.~Criminisi, ``Training
  cnns with low-rank filters for efficient image classification,'' {\em arXiv
  preprint arXiv:1511.06744}, 2015.

\bibitem{hubara2016quantized}
I.~Hubara, M.~Courbariaux, D.~Soudry, R.~El-Yaniv, and Y.~Bengio, ``Quantized
  neural networks: Training neural networks with low precision weights and
  activations,'' {\em arXiv preprint arXiv:1609.07061}, 2016.

\bibitem{gysel2016hardware}
P.~Gysel, M.~Motamedi, and S.~Ghiasi, ``Hardware-oriented approximation of
  convolutional neural networks,'' {\em arXiv preprint arXiv:1604.03168}, 2016.

\bibitem{zhou2017balanced}
S.-C. Zhou, Y.-Z. Wang, H.~Wen, Q.-Y. He, and Y.-H. Zou, ``Balanced
  quantization: An effective and efficient approach to quantized neural
  networks,'' {\em Journal of Computer Science and Technology}, vol.~32, no.~4,
  pp.~667--682, 2017.

\bibitem{denil2013predicting}
M.~Denil, B.~Shakibi, L.~Dinh, N.~de~Freitas, {\em et~al.}, ``Predicting
  parameters in deep learning,'' in {\em Advances in Neural Information
  Processing Systems}, pp.~2148--2156, 2013.

\bibitem{novikov2015tensorizing}
A.~Novikov, D.~Podoprikhin, A.~Osokin, and D.~P. Vetrov, ``Tensorizing neural
  networks,'' in {\em Advances in Neural Information Processing Systems},
  pp.~442--450, 2015.

\bibitem{lin2016towards}
S.~Lin, R.~Ji, X.~Guo, X.~Li, {\em et~al.}, ``Towards convolutional neural
  networks compression via global error reconstruction.,'' in {\em IJCAI},
  pp.~1753--1759, 2016.

\bibitem{lebedev2014speeding}
V.~Lebedev, Y.~Ganin, M.~Rakhuba, I.~Oseledets, and V.~Lempitsky, ``Speeding-up
  convolutional neural networks using fine-tuned cp-decomposition,'' {\em arXiv
  preprint arXiv:1412.6553}, 2014.

\bibitem{kim2015compression}
Y.-D. Kim, E.~Park, S.~Yoo, T.~Choi, L.~Yang, and D.~Shin, ``Compression of
  deep convolutional neural networks for fast and low power mobile
  applications,'' {\em arXiv preprint arXiv:1511.06530}, 2015.

\bibitem{lin2013network}
M.~Lin, Q.~Chen, and S.~Yan, ``Network in network,'' {\em arXiv preprint
  arXiv:1312.4400}, 2013.

\bibitem{li2014multilinear}
Q.~Li and D.~Schonfeld, ``Multilinear discriminant analysis for higher-order
  tensor data classification,'' {\em IEEE transactions on pattern analysis and
  machine intelligence}, vol.~36, no.~12, pp.~2524--2537, 2014.

\bibitem{yan2005discriminant}
S.~Yan, D.~Xu, Q.~Yang, L.~Zhang, X.~Tang, and H.-J. Zhang, ``Discriminant
  analysis with tensor representation,'' in {\em Computer Vision and Pattern
  Recognition, 2005. CVPR 2005. IEEE Computer Society Conference on}, vol.~1,
  pp.~526--532, IEEE, 2005.

\bibitem{zhou2013tensor}
H.~Zhou, L.~Li, and H.~Zhu, ``Tensor regression with applications in
  neuroimaging data analysis,'' {\em Journal of the American Statistical
  Association}, vol.~108, no.~502, pp.~540--552, 2013.

\bibitem{thanh2017tensor}
D.~T. Thanh, J.~Kanniainen, M.~Gabbouj, and A.~Iosifidis, ``Tensor
  representation in high-frequency financial data for price change
  prediction,'' {\em arXiv preprint arXiv:1709.01268}, 2017.

\bibitem{tao2007general}
D.~Tao, X.~Li, X.~Wu, and S.~J. Maybank, ``General tensor discriminant analysis
  and gabor features for gait recognition,'' {\em IEEE Transactions on Pattern
  Analysis and Machine Intelligence}, vol.~29, no.~10, 2007.

\bibitem{lu2008mpca}
H.~Lu, K.~N. Plataniotis, and A.~N. Venetsanopoulos, ``Mpca: Multilinear
  principal component analysis of tensor objects,'' {\em IEEE Transactions on
  Neural Networks}, vol.~19, no.~1, pp.~18--39, 2008.

\bibitem{guo2012tensor}
W.~Guo, I.~Kotsia, and I.~Patras, ``Tensor learning for regression,'' {\em IEEE
  Transactions on Image Processing}, vol.~21, no.~2, pp.~816--827, 2012.

\bibitem{zhang2017shufflenet}
X.~Zhang, X.~Zhou, M.~Lin, and J.~Sun, ``Shufflenet: An extremely efficient
  convolutional neural network for mobile devices,'' {\em arXiv preprint
  arXiv:1707.01083}, 2017.

\bibitem{wangdesign}
M.~Wang, B.~Liu, and H.~Foroosh, ``Design of efficient convolutional layers
  using single intra-channel convolution, topological subdivisioning and
  spatial “bottleneck” structure,''

\bibitem{szegedy2017inception}
C.~Szegedy, S.~Ioffe, V.~Vanhoucke, and A.~A. Alemi, ``Inception-v4,
  inception-resnet and the impact of residual connections on learning.,'' in
  {\em AAAI}, pp.~4278--4284, 2017.

\bibitem{howard2017mobilenets}
A.~G. Howard, M.~Zhu, B.~Chen, D.~Kalenichenko, W.~Wang, T.~Weyand,
  M.~Andreetto, and H.~Adam, ``Mobilenets: Efficient convolutional neural
  networks for mobile vision applications,'' {\em arXiv preprint
  arXiv:1704.04861}, 2017.

\bibitem{ioffe2015batch}
S.~Ioffe and C.~Szegedy, ``Batch normalization: Accelerating deep network
  training by reducing internal covariate shift,'' in {\em International
  Conference on Machine Learning}, pp.~448--456, 2015.

\bibitem{krizhevsky2012imagenet}
A.~Krizhevsky, I.~Sutskever, and G.~E. Hinton, ``Imagenet classification with
  deep convolutional neural networks,'' in {\em Advances in neural information
  processing systems}, pp.~1097--1105, 2012.

\bibitem{simonyan2014very}
K.~Simonyan and A.~Zisserman, ``Very deep convolutional networks for
  large-scale image recognition,'' {\em arXiv preprint arXiv:1409.1556}, 2014.

\bibitem{kossaifi2017tensor}
J.~Kossaifi, A.~Khanna, Z.~C. Lipton, T.~Furlanello, and A.~Anandkumar,
  ``Tensor contraction layers for parsimonious deep nets,'' {\em arXiv preprint
  arXiv:1706.00439}, 2017.

\bibitem{kolda2009tensor}
T.~G. Kolda and B.~W. Bader, ``Tensor decompositions and applications,'' {\em
  SIAM review}, vol.~51, no.~3, pp.~455--500, 2009.

\bibitem{ronneberger2015u}
O.~Ronneberger, P.~Fischer, and T.~Brox, ``U-net: Convolutional networks for
  biomedical image segmentation,'' in {\em International Conference on Medical
  Image Computing and Computer-Assisted Intervention}, pp.~234--241, Springer,
  2015.

\bibitem{maas2013rectifier}
A.~L. Maas, A.~Y. Hannun, and A.~Y. Ng, ``Rectifier nonlinearities improve
  neural network acoustic models,'' in {\em Proc. ICML}, vol.~30, 2013.

\bibitem{springenberg2014striving}
J.~T. Springenberg, A.~Dosovitskiy, T.~Brox, and M.~Riedmiller, ``Striving for
  simplicity: The all convolutional net,'' {\em arXiv preprint
  arXiv:1412.6806}, 2014.

\bibitem{xu2015empirical}
B.~Xu, N.~Wang, T.~Chen, and M.~Li, ``Empirical evaluation of rectified
  activations in convolutional network,'' {\em arXiv preprint
  arXiv:1505.00853}, 2015.

\bibitem{krizhevsky2009learning}
A.~Krizhevsky and G.~Hinton, ``Learning multiple layers of features from tiny
  images,'' 2009.

\bibitem{netzer2011reading}
Y.~Netzer, T.~Wang, A.~Coates, A.~Bissacco, B.~Wu, and A.~Y. Ng, ``Reading
  digits in natural images with unsupervised feature learning,'' in {\em NIPS
  workshop on deep learning and unsupervised feature learning}, vol.~2011,
  p.~5, 2011.

\bibitem{rumelhart1988learning}
D.~E. Rumelhart, G.~E. Hinton, R.~J. Williams, {\em et~al.}, ``Learning
  representations by back-propagating errors,'' {\em Cognitive modeling},
  vol.~5, no.~3, p.~1, 1988.

\bibitem{kingma2014adam}
D.~Kingma and J.~Ba, ``Adam: A method for stochastic optimization,'' {\em arXiv
  preprint arXiv:1412.6980}, 2014.

\bibitem{srivastava2014dropout}
N.~Srivastava, G.~E. Hinton, A.~Krizhevsky, I.~Sutskever, and R.~Salakhutdinov,
  ``Dropout: a simple way to prevent neural networks from overfitting.,'' {\em
  Journal of machine learning research}, vol.~15, no.~1, pp.~1929--1958, 2014.

\bibitem{he2015delving}
K.~He, X.~Zhang, S.~Ren, and J.~Sun, ``Delving deep into rectifiers: Surpassing
  human-level performance on imagenet classification,'' in {\em Proceedings of
  the IEEE international conference on computer vision}, pp.~1026--1034, 2015.

\bibitem{chollet2015}
F.~Chollet, ``keras.'' https://github.com/fchollet/keras, 2015.

\bibitem{tensorflow2015paper}
M.~Abadi, A.~Agarwal, P.~Barham, E.~Brevdo, Z.~Chen, C.~Citro, G.~S. Corrado,
  A.~Davis, J.~Dean, M.~Devin, S.~Ghemawat, I.~Goodfellow, A.~Harp, G.~Irving,
  M.~Isard, Y.~Jia, R.~Jozefowicz, L.~Kaiser, M.~Kudlur, J.~Levenberg,
  D.~Man\'{e}, R.~Monga, S.~Moore, D.~Murray, C.~Olah, M.~Schuster, J.~Shlens,
  B.~Steiner, I.~Sutskever, K.~Talwar, P.~Tucker, V.~Vanhoucke, V.~Vasudevan,
  F.~Vi\'{e}gas, O.~Vinyals, P.~Warden, M.~Wattenberg, M.~Wicke, Y.~Yu, and
  X.~Zheng, ``{TensorFlow}: Large-scale machine learning on heterogeneous
  systems,'' 2015.
\newblock Software available from tensorflow.org.

\end{thebibliography}
\bibliographystyle{ieeetr}

\end{document}